\documentclass[11pt,letterpaper]{article}

\usepackage[letterpaper,margin=1in]{geometry}
\usepackage[T1]{fontenc}
\usepackage[utf8]{inputenc}
\usepackage{lmodern}
\usepackage[table]{xcolor}
\usepackage{graphicx}
\usepackage{booktabs}
\usepackage{amsmath,amssymb}
\usepackage{pifont}
\usepackage{multirow}
\usepackage{tabularx}
\usepackage[round,authoryear]{natbib}
\setcitestyle{authoryear,round,aysep={ }}
\let\cite\citep
\usepackage{caption}
\usepackage[hyphens]{url}
\usepackage[hidelinks]{hyperref}

\definecolor{headerblue}{RGB}{225,235,245}

\title{FlowForm: Synergizing Fluid Physics with Topological Consistency for Satellite Flood Synthesis}

\author{%
\normalsize Weihui Zhang \quad
Ruizhi Wang \quad
Hongye Xu\\[0.4em]
\normalsize Huiqiong Wang \quad
Li Sun \quad
Mingli Song\\[0.55em]
\small Zhejiang University, Zhejiang, China\\[0.35em]
\footnotesize
\href{mailto:weihuizhang@zju.edu.cn}{weihuizhang@zju.edu.cn} \quad
\href{mailto:ruizhiwang@zju.edu.cn}{ruizhiwang@zju.edu.cn} \quad
\href{mailto:hongyexu@zju.edu.cn}{hongyexu@zju.edu.cn}\\[-0.05em]
\footnotesize
\href{mailto:huiqiong_wang@zju.edu.cn}{huiqiong\_wang@zju.edu.cn} \quad
\href{mailto:lsun@zju.edu.cn}{lsun@zju.edu.cn} \quad
\href{mailto:brooksong@zju.edu.cn}{brooksong@zju.edu.cn}}
\date{}

\hypersetup{
  pdftitle={FlowForm: Synergizing Fluid Physics with Topological Consistency for Satellite Flood Synthesis},
  pdfauthor={Weihui Zhang, Ruizhi Wang, Hongye Xu, Huiqiong Wang, Li Sun, Mingli Song},
  pdfsubject={Satellite flood synthesis with physics-inspired latent regularization and structure-aware conditioning},
  pdfkeywords={satellite imagery, flood synthesis, diffusion models, physics-informed learning}
}

\begin{document}

\maketitle

\begin{abstract}

Developing robust flood assessment models requires high-quality paired satellite imagery, yet such data remain scarce for flood-specific image generation. Although generative models provide a promising means of data augmentation, existing methods often yield implausible spatial layouts of flooded regions and distort scene structures. We propose FlowForm, a framework for satellite flood synthesis that integrates SWE-inspired latent regularization with structure-aware conditioning. The Flood Descriptor Module (FDM) imposes differentiable penalties on residuals of the steady-state Shallow Water Equation in auxiliary latent fields at the diffusion bottleneck. The Terrain Anchor Adapter (TAA) injects depth, semantic, and edge features at four encoder scales of the U-Net. We further curate FloodScape, a large-scale, high-resolution dataset comprising paired satellite images acquired before and after disasters. In addition to standard image-generation metrics, we evaluate the consistency of flooded regions, zero-shot generalization to a geographically held-out flood event, and sensitivity to individual components. Across all reported comparisons, FlowForm achieves higher visual fidelity, greater similarity between paired images, and stronger consistency of flooded regions.

\end{abstract}

\section{Introduction}

As rainstorm-induced flood risks increase under warming~\cite{zhang2022reconciling}, timely post-disaster assessment from satellite imagery can support humanitarian assistance and disaster recovery~\cite{gupta2019xbd}. Data-driven post-disaster and change-detection models have advanced automated assessment~\cite{asad2023natural, zang2025changediff, zhang2023new, safavi2022comparative}; however, the reliability of these models depends on representative training data. Unlike ordinary imagery, high-quality observations of disasters remain scarce in remote sensing~\cite{9460988, gupta2019xbd, wang2025ebd}. This limitation is particularly acute for change analysis, which must distinguish changes related to the event from persistent scenes using spatially aligned pre-disaster and post-disaster observations. Therefore, a useful training pair requires more than simply obtaining two acquisitions from the same location: the images must be sufficiently co-registered, the flood must be visible, and unrelated changes must be minimal. Cloud cover and the short observation windows of flood events make it difficult to collect such pairs at scale.

Generative models offer a scalable approach to data augmentation~\cite{goodfellow2020generative, ho2020denoising, song2020denoising, rombach2022high, tang2025aerogen, toker2024satsynth}. Diffusion models are particularly effective in this context, as they synthesize detailed imagery while accommodating text, metadata, or image conditioning. Prior studies in remote sensing have synthesized post-disaster scenes from pre-disaster inputs, demonstrating the feasibility of this data-centric paradigm~\cite{khanna2023diffusionsat, lutjens2024generating, rui2021disastergan}. However, a visually plausible satellite image does not necessarily represent a logically valid transformation of the provided pre-event scene. A generative model may produce convincing water textures while placing inundations in incorrect regions, or it may depict a generally credible post-disaster scene while arbitrarily altering buildings, roads, and vegetation that should remain unchanged. The synthesis of flood imagery must therefore explicitly control both event-specific changes and the content expected to remain invariant.

\begin{figure}[htbp]
    \centering
    \includegraphics[width=1\linewidth]{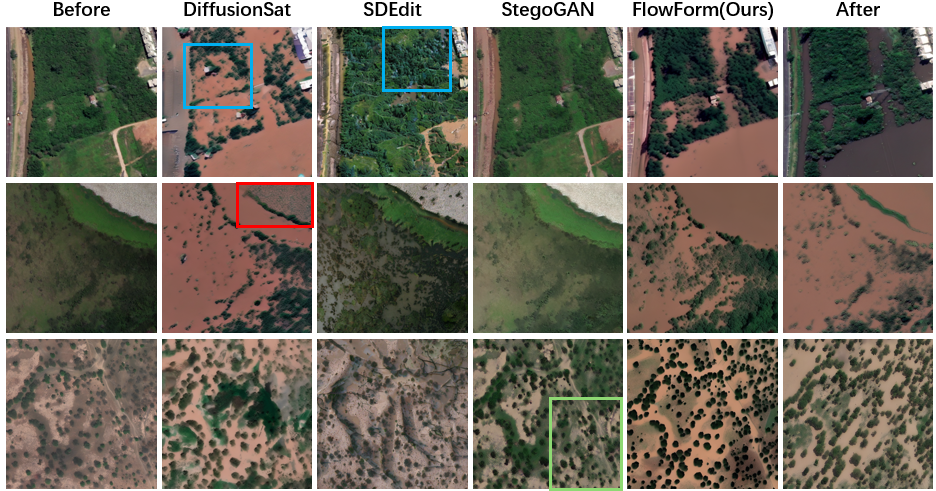}

    \caption{Comparison of structural, semantic, and flood-region consistency. Blue, red, and green boxes indicate geometric distortion, semantic shift, and fragmented inundation appearance, respectively. FlowForm better retains scene structure while producing more coherent flood regions in the displayed examples.}
    \label{fig:structural_semantic_consistency}
\end{figure}

The primary challenge lies in the localization of flood regions. Data-driven generators learn the appearance of water through RGB correlations; however, an RGB reconstruction objective fails to explicitly encode the relationship between the extent of inundation and local terrain gradients, land-cover categories, or connected water bodies. Consequently, the generated water texture may appear visually credible even if the predicted region is fragmented or misaligned with the surrounding scene. The green boxes in Fig.~\ref{fig:structural_semantic_consistency} illustrate instances of disconnected or misplaced inundation appearances. This limitation highlights the necessity for a spatial training signal that operates directly on flood-state features rather than solely on RGB textures.

The second challenge is to preserve persistent scene content. Buildings, roads, vegetation boundaries, and other non-flood structures serve as stable references for interpreting disaster-induced changes. Generic image translation objectives do not explicitly disentangle these elements from flood-related appearance changes. Consequently, the model may displace boundaries, deform infrastructure, or alter local semantic classes when synthesizing flooded regions. Such errors are particularly detrimental in paired image generation because they confound actual disaster-induced changes with artifacts introduced by the generator. The blue and red boxes in Fig.~\ref{fig:structural_semantic_consistency} illustrate examples of geometric distortion and semantic shift, respectively. Structural conditions derived from the pre-event observation can constrain the geometry and semantics of persistent content while allowing flood-affected regions to be modified.

Dataset construction presents a related practical challenge. Public paired flood datasets are limited in scale and fidelity~\cite{khanna2023diffusionsat, gupta2019xbd}, and nominal pre/post labels do not inherently guarantee a useful synthesis pair. Temporal misalignment, limited flood visibility, missing observations, and cloud contamination diminish the effective diversity of available data. Furthermore, these factors compromise evaluation reliability because a generated image may be compared against a target whose visual changes are ambiguous. Consequently, a flood-specific benchmark should explicitly verify pair validity while ensuring geographic and event diversity.

To address these requirements, FlowForm combines SWE-inspired latent regularization with structure-aware conditioning. Specifically, the Flood Descriptor Module (FDM) predicts a latent depth proxy alongside auxiliary flow fields from the diffusion bottleneck. Finite-difference operators evaluate simplified steady-state SWE residuals over these fields, and add the resulting squared residuals to the training objective. In parallel, the Terrain Anchor Adapter (TAA) adaptively selects among relative-depth, semantic, and edge features based on the diffusion timestep and spatial location, and injects the resulting structural representation into four scales of the U-Net encoder. While the FDM operates on flood-state features, the TAA conveys persistent scene cues through the visual backbone. FloodScape provides the aligned pre-event and post-event imagery and structural conditions required to train and evaluate the proposed formulation.

The contributions can be summarized as follows:
\begin{itemize}
    \item
        \textbf{Flood Descriptor Module (FDM):}

        We design an auxiliary branch at the U-Net bottleneck to predict a latent depth proxy and auxiliary flow fields. Steady-state SWE residuals computed from these predictions serve as an additional regularization term during training.
    \item
        \textbf{Terrain Anchor Adapter (TAA):}

        TAA injects relative depth, semantic, and edge features conditioned on the diffusion timestep, providing structure-aware guidance that preserves persistent scene content during flood synthesis.
    \item
        \textbf{FloodScape Dataset:}

        FloodScape comprises approximately 10,000 high-resolution pre- and post-disaster satellite image pairs, together with spatially aligned structural conditions for training and evaluating flood synthesis methods.
\end{itemize}

\section{Related Work}
\label{sec:related_work}

\subsection{Remote Sensing Datasets for Flood Disasters}

High-quality datasets underpin robust remote sensing applications, ranging from traditional land cover classification to critical tasks such as disaster management and change detection~\cite{zhu2025skysense, bai2020pyramid, zang2025changediff}. Although general-purpose Earth observation datasets offer large-scale image archives, they often fail to satisfy the specialized requirements of disaster-related tasks. Applications such as change detection, post-disaster assessment, and generative disaster simulation inherently require accurately aligned pre- and post-disaster image pairs. Existing disaster-specific datasets, including xBD~\cite{gupta2019xbd} and EBD~\cite{wang2025ebd}, provide multi-temporal imagery but still exhibit notable limitations. Specifically, a substantial portion of the imagery labeled as post-disaster lacks visible evidence of disaster-induced changes. Furthermore, persistent challenges such as dense cloud cover and widespread data gaps substantially reduce the availability of valid image pairs. This scarcity of high-quality, temporally aligned pre- and post-disaster samples hinders the training of generative models for post-disaster scene reconstruction. To address this limitation, we present a rigorously curated flood-specific dataset containing precisely paired pre- and post-disaster satellite images, explicitly designed to support generative modeling in this domain.

\subsection{Diffusion Models in Remote Sensing}

Diffusion Probabilistic Models (DPMs)~\cite{ho2020denoising} have emerged as a dominant generative paradigm in computer vision, offering greater training stability and sample diversity than Generative Adversarial Networks (GANs)~\cite{goodfellow2020generative}. In remote sensing, recent methods such as Text2Earth~\cite{liu2025text2earth} and Crs-diff~\cite{tang2024crs} have successfully applied diffusion models to satellite image synthesis, whereas other studies~\cite{khanna2023diffusionsat, lutjens2024generating} have begun exploring the generation of flood-affected scenes. However, existing generative approaches to flood scenarios are typically formulated as image-to-image style transfer systems that overlook the topological continuity of water bodies. Moreover, they rely primarily on raw optical imagery without incorporating essential geometric and semantic constraints, such as depth maps and semantic masks. The absence of such constraints often leads to structural artifacts and semantic inconsistencies, including distorted infrastructure and misaligned scene elements. To address these limitations, FlowForm introduces the Terrain Anchor Adapter, which integrates multimodal spatial and structural priors into U-Net features in a diffusion-timestep-dependent manner. By enforcing geometric consistency between the generated content and the underlying terrain, this mechanism improves the physical plausibility and structural fidelity of the generated post-disaster scenes.

\begin{figure*}[!t]
    \centering
    \includegraphics[width=0.8\textwidth]{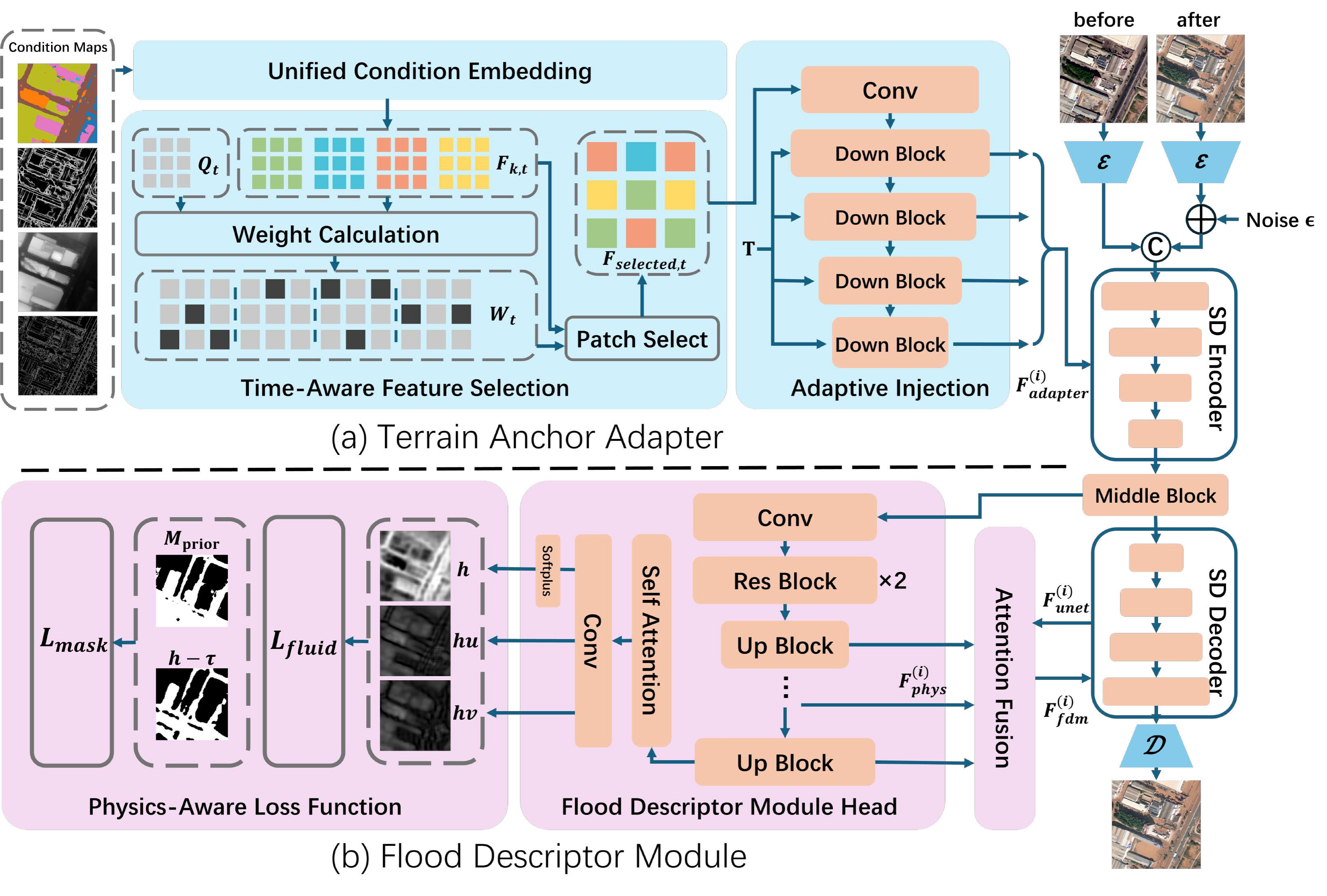}

    \caption{Overall FlowForm architecture: (a) Terrain Anchor Adapter (TAA) and (b) Flood Descriptor Module (FDM).}
    \label{fig:overall_architecture}
\end{figure*}

\subsection{Physical Consistency in Image Generation}

In disaster simulation, physical plausibility is as critical as visual realism. Physics-informed neural networks (PINNs)~\cite{raissi2019physics} have demonstrated that physical laws, such as the shallow water equations (SWE)~\cite{qi2024physics} and other partial differential equations (PDEs)~\cite{harandi2024mixed, cuomo2022scientific}, can be incorporated into loss functions to enforce the conservation of mass and momentum. However, a significant gap remains between these physics-based formulations and the requirements of data-driven visual generation. Diffusion models, such as Stable Diffusion, effectively capture pixel-level statistical distributions, yet they do not explicitly encode physical constraints. Conversely, PINN-based approaches typically regress scalar fields, such as water depth and velocity, rather than generating photorealistic RGB imagery. Consequently, their optimization objectives are difficult to reconcile: diffusion models prioritize visual fidelity, whereas physics-based methods must satisfy strict governing equations. To bridge this gap, FlowForm introduces the Flood Descriptor Module, which integrates fluid dynamics constraints directly into the visual generation pipeline. By enforcing physical consistency during synthesis, FlowForm generates flood scenarios that combine photorealistic quality with adherence to fundamental hydrodynamic principles.

\section{Method}

FlowForm adopts Stable Diffusion 2.1~\cite{rombach2022high} as its backbone and augments it with two core components: a Terrain Anchor Adapter (TAA), which provides static structural anchors to preserve the underlying topology, and a Flood Descriptor Module (FDM), which applies simplified steady-state SWE residuals to auxiliary latent fields. The overall architecture is depicted in Fig.~\ref{fig:overall_architecture}.

\subsection{Flood Descriptor Module}

FDM takes U-Net features as input and predicts a latent depth proxy $h$ and auxiliary flow fields $(hu, hv)$. Differentiable finite-difference operators evaluate simplified steady-state SWE residuals, whose squared values enter the training objective. The residuals penalize deviations from the mass and momentum equations, including terms coupled to local terrain gradients.

\subsubsection{Flood Descriptor Module Head}

The FDM Head serves as an auxiliary neural branch designed to decode latent flood-state variables from the high-level semantic features of the backbone. Located immediately after the mid-block of the U-Net, this module processes the bottleneck features through a sequence of residual blocks and upsampling layers that mirror the resolution hierarchy of the U-Net decoder. This branch performs two functions:

\begin{itemize}
    \item \textbf{Residual-Guided Feature Extraction:}

    During the upsampling process, intermediate feature maps $\{F_{fdm}^{(i)}\}$ are extracted at multiple scales. These features are spatially aligned with the U-Net decoder and retained to guide visual synthesis through the fusion module.

    \item \textbf{State Prediction:}

    The final layer reconstructs a nonnegative latent depth proxy $h$ and auxiliary flow fields $(q_x, q_y)$, denoted as $(hu, hv)$ for consistency with SWE notation. We apply a Softplus activation to the $h$ channel to impose non-negativity.
\end{itemize}

The auxiliary branch is jointly optimized via the diffusion reconstruction objective and the SWE-inspired regularizers defined below.

\begin{figure}[!t]
    \centering
    \includegraphics[width=0.8\columnwidth]{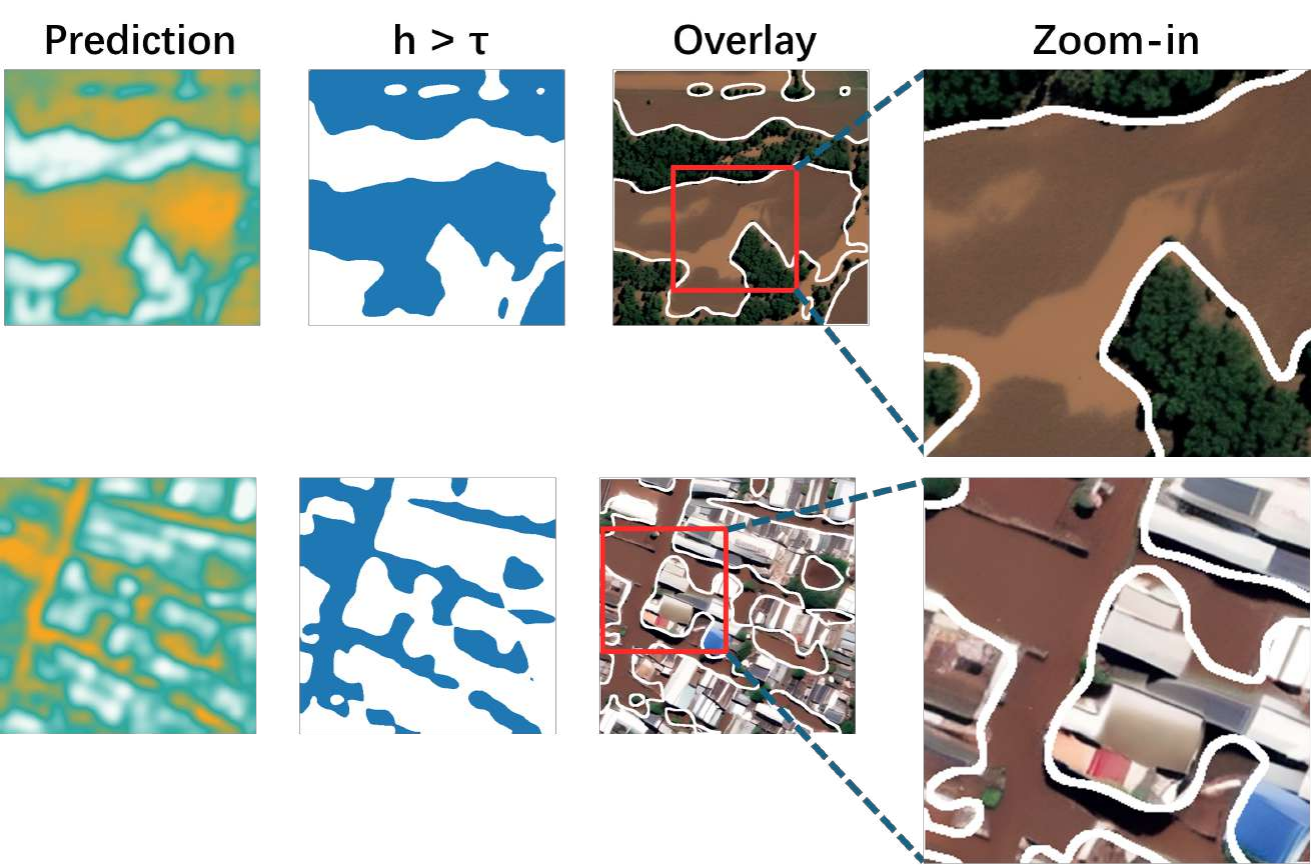}

    \caption{Visualization of the latent depth proxy $h$. Each row shows the continuous prediction, thresholded $h>\tau$ mask, boundary overlay, and linked zoom-in view.}
    \label{fig:h_state_visualization}
\end{figure}

Fig.~\ref{fig:h_state_visualization} illustrates the latent depth proxy $h$, the corresponding thresholded flood mask $h > \tau$, and the mask boundary overlaid on the generated image. The linked magnified views show how the inferred boundary follows the local inundation structure and make boundary estimation errors readily observable.

\subsubsection{SWE-Inspired Latent Regularization}

The composite regularization loss contains an SWE residual term $\mathcal{L}_{fluid}$ and a semantic guidance term $\mathcal{L}_{mask}$.

\paragraph{Steady-State SWE Residual ($\mathcal{L}_{fluid}$)}

We use a simplified steady-state form of the Shallow Water Equations (SWE) as SWE-inspired regularization for the auxiliary flow fields. For static post-disaster image synthesis, we set time derivatives to zero and compute mass-equation ($r_{m}$) and momentum-equation ($r_{u}, r_{v}$) residuals:

The mass-equation residual ($r_{m}$) is defined as:
\begin{equation}
    r_{m} = \frac{\partial (hu)}{\partial x} + \frac{\partial (hv)}{\partial y}
\end{equation}

The momentum-equation residuals ($r_{u}$ and $r_{v}$) are defined as:
\begin{equation}
    \begin{aligned}
        r_{u} &= \frac{\partial (hu^2 + \frac{1}{2}gh^2)}{\partial x} + \frac{\partial (huv)}{\partial y} + gh \frac{\partial z}{\partial x} \\
        r_{v} &= \frac{\partial (huv)}{\partial x} + \frac{\partial (hv^2 + \frac{1}{2}gh^2)}{\partial y} + gh \frac{\partial z}{\partial y}
    \end{aligned}
\end{equation}

where $g$ denotes the gravitational constant used in the surrogate residual, and $z$ is a relative-height proxy derived from monocular depth estimation. The proxy supplies local terrain ordering and gradients to the regularizer.

Directly summing the mass and momentum residuals causes training instability due to the substantial differences in their magnitudes. We therefore divide each residual term by its batch-wise mean magnitude (e.g., $\bar{r}_m$). The resulting SWE residual loss is:

\begin{equation}
    \mathcal{L}_{fluid} = \left\| \frac{r_{m}}{\bar{r}_m} \right\|^{2} + \left\| \frac{r_{u}}{\bar{r}_u} \right\|^{2} + \left\| \frac{r_{v}}{\bar{r}_v} \right\|^{2}
\end{equation}

\paragraph{Semantic Initialization via Masking ($\mathcal{L}_{mask}$)}

During the early stages of training, the FDM may converge to the trivial all-zero solution ($h=0, hu=0, hv=0$), which yields zero residuals but provides no useful spatial signal. A semantic flood-candidate prior supplies non-zero spatial targets to the auxiliary branch.

We construct $M_{prior}$ from semantic segmentation by assigning positive targets to flood-candidate classes (e.g., roads, fields, and water bodies) and negative targets to likely obstacles (e.g., buildings and trees). Binary cross-entropy applies these targets to $h$:
\begin{equation}
    \mathcal{L}_{mask} = \text{BCE}(\text{Sigmoid}(h - \tau), M_{prior})
\end{equation}

where $\tau$ is a threshold on the latent proxy $h$. Positive mask targets provide a non-zero training signal for the auxiliary branch.

\subsubsection{Cross-Attention Fusion}

A cross-attention block fuses the residual-regularized features before each U-Net decoder upsampling block. At decoder level $i$, visual features $F_{unet}^{(i)}$ form the queries ($Q$), while FDM features $F_{fdm}^{(i)}$ form the keys ($K$) and values ($V$). The fused feature $F_{fused}^{(i)}$ is:
\begin{equation}
    F_{fused}^{(i)} = W_{out}\left( \text{Softmax}\left(\beta \cdot Q K^{T} \right) V \right)
\end{equation}

where $\beta$ denotes a learnable temperature parameter that controls the sharpness of the attention map, and $W_{out}$ represents an output linear projection matrix to restore the feature dimensions.
The attention output mixes latent flood-state features into the visual decoder before upsampling.

\subsection{Terrain Anchor Adapter}

The Terrain Anchor Adapter (TAA) injects structural priors into the U-Net encoder to retain persistent scene features. Although dual-branch architectures such as ControlNet~\cite{zhang2023adding} excel in reference-based conditional generation, their substantial parameter overhead makes them impractical for efficient joint training in this specific setting. Unlike computationally intensive models or autoregressive approaches requiring complex sequential modeling~\cite{pan2025pixelponder}, the TAA operates as a lightweight, parallelizable module. At each timestep, it selects a structural prior for every spatial location before multi-scale U-Net injection.

\subsubsection{Unified Condition Embedding}

We use the mixture-of-experts strategy from UniControl~\cite{qin2023unicontrol} to align heterogeneous conditions (e.g., depth, semantics, Canny, and HED) within a shared feature space. Following PixelPonder~\cite{pan2025pixelponder}, the aligned features are flattened into patch tokens and combined with rotary positional embeddings (RoPE).

\subsubsection{Time-Aware Feature Selection}

TAA selects among structural signals separately at each diffusion stage and spatial location.
The network converts multi-modal conditions into time-conditioned features $\hat{F}_{k,t}$ and spatial query features $Q_t$. Their matching logits $E_{n,k,t}$ are $L_2$-normalized across the $K$ conditions and passed through a softmax, producing a relevance map $W_t \in \mathbb{R}^{N \times K}$:
\begin{equation}
W_{n, \cdot, t} = \text{softmax}\left( \frac{E_{n, \cdot, t}}{\| E_{n, \cdot, t} \|_2} \right)
\end{equation}

where, $E_{n, k, t}$ denotes the pre-computed matching logit for the $k$-th condition at spatial location $n$ and timestep $t$.

At each spatial location, an argmax over $W_t$ selects one feature patch from the original inputs:
\begin{equation}
F_{selected, t}^{(n)} = \sum_{k=1}^{K} \mathbb{I}\!\left(k = \operatorname*{arg\,max}_{m \in \{1, \dots, K\}} W_{n,m,t} \right) F_k^{(n)}
\end{equation}

We use the straight-through estimator (STE) to propagate gradients through the discrete selection operation. The scores $W_t$ vary with $t$, while the selected payload carries the original structural-prior features.

\subsubsection{Adaptive Injection}

Four adapter blocks transform $F_{selected}$ into conditioned representations $\{F_{adapter}^{(i)}\}_{i=1}^4$ at four resolution scales. Side connections inject each representation into the corresponding U-Net encoder level. Before each adapter block, the timestep embedding $t_{emb}$ is added to its input. Let $x^{(i-1)}$ denote the input to the $i$-th block, with $x^{(0)}=F_{selected}$. The time-modulated features are:
\begin{equation}
x^{(i)} = \text{AdapterBlock}_i\left( x^{(i-1)} + t_{emb} \right), \quad F_{adapter}^{(i)} = x^{(i)}
\end{equation}

Thus, each selected structural prior enters all four encoder scales after timestep modulation.

\section{FloodScape Dataset}
\label{sec:dataset}
\subsection{Data Acquisition and Preprocessing}

The proposed dataset is mainly derived from the Maxar Open Data Program~\cite{maxar_open_data}, supplemented by samples from the xBD~\cite{gupta2019xbd} and eBD~\cite{wang2025ebd} datasets. We apply quality control across all sources and discard samples with severe cloud cover, large invalid regions, or no clearly observable flooding in the post-event imagery. This curation yields FloodScape, a collection of approximately 10,000 high-quality, spatially aligned pre- and post-disaster satellite image pairs.

\subsection{Multi-Modal Condition Construction}

To provide complementary terrain, semantic, and structural priors, we generate four spatially aligned condition maps from each pre-disaster image. Specifically, Depth-Anything-V2~\cite{yang2024depth} and SkySense-O~\cite{zhu2025skysense} are used to obtain depth and semantic segmentation maps, while the Canny operator~\cite{canny2009computational} and HED model~\cite{xie2015holistically} extract local edges and structural boundaries, respectively.

\begin{table*}[!t]
    \centering
    \footnotesize
    \setlength{\tabcolsep}{0.8pt}
    \begin{tabular*}{\textwidth}{@{\extracolsep{\fill}}lccccccc@{}}
        \toprule
        & \multicolumn{5}{c}{\textbf{Image Quality}} & \multicolumn{2}{c}{\textbf{Flood-region Consistency}} \\
        \cmidrule(lr){2-6}\cmidrule(lr){7-8}
        Method & FID $\downarrow$ & SSIM $\uparrow$ & LPIPS $\downarrow$ & CLIP $\uparrow$ & PSNR $\uparrow$ & IoU $\uparrow$ & FVPS $\uparrow$ \\
        \midrule
        \rowcolor{headerblue}\multicolumn{8}{c}{\textbf{GAN-based Methods}} \\
        CycleGAN~\cite{CycleGAN} & 81.2357 & 0.4233 & 0.6676 & 0.8346 & 15.4788 & 0.2175 & 0.2629 \\
        CUT~\cite{CUT} & 76.6915 & 0.4293 & 0.6727 & 0.8317 & 15.0707 & 0.2212 & 0.2640 \\
        StegoGAN~\cite{StegoGAN} & 90.6001 & 0.3780 & 0.6762 & 0.8224 & 15.6771 & 0.1938 & 0.2425 \\
        EnCo~\cite{EnCo} & 87.0716 & 0.4469 & 0.6895 & 0.8244 & \underline{15.7422} & 0.1762 & 0.2248 \\
        \rowcolor{headerblue}\multicolumn{8}{c}{\textbf{Diffusion-based Methods}} \\
        Pix2Pix-Zero~\cite{Pix2Pix-Zero} & 153.8684 & 0.4379 & 0.6924 & 0.7286 & 14.6958 & 0.2192 & 0.2560 \\
        CycleNet~\cite{xu2023cyclenet} & 99.3221 & 0.3820 & 0.6742 & 0.8260 & 14.7848 & 0.2069 & 0.2531 \\
        SDEdit~\cite{SDEdit} & 104.9919 & 0.3484 & 0.6950 & 0.7818 & 14.6450 & 0.2147 & 0.2520 \\
        CycleDiffusion~\cite{CycleDiffusion} & 154.1718 & 0.4243 & 0.6494 & 0.8334 & 15.6980 & 0.1932 & 0.2491 \\
        ILVR~\cite{ILVR} & 173.4569 & 0.2552 & 0.7078 & 0.7945 & 15.3755 & 0.1113 & 0.1612 \\
        EGSDE~\cite{EGSDE} & 87.6277 & 0.4405 & 0.6577 & 0.8334 & 15.2813 & 0.1227 & 0.1806 \\
        DiffusionSat~\cite{khanna2023diffusionsat} & \underline{76.3459} & \underline{0.4849} & \underline{0.6062} & \underline{0.8552} & 15.6403 & \underline{0.3818} & \underline{0.3877} \\
        \midrule
        \textbf{FlowForm} & \textbf{71.7999} & \textbf{0.4941} & \textbf{0.5671} & \textbf{0.8691} & \textbf{15.7556} & \textbf{0.4413} & \textbf{0.4371} \\
        \bottomrule
    \end{tabular*}

    \caption{Quantitative comparison on FloodScape. The best results are in \textbf{bold}, and the second-best results are \underline{underlined}.}
    \label{tab:sota_comparison}
\end{table*}

\section{Experiments}
\subsection{Experimental Settings}
\subsubsection{Dataset Preparation.}

We evaluate FlowForm on the FloodScape dataset, which is divided into three subsets: 8,716 image pairs for training, 969 for testing, and 442 for zero-shot generalization. The zero-shot set consists of data from the 2022 South Africa flood event, a geographic region explicitly excluded from training, enabling evaluation on an unseen event. All images are resized to $512 \times 512$ during both training and evaluation to standardize the input resolution and ensure a fair comparison across different methods.

\subsubsection{Baselines.}

We compare FlowForm with GAN-based baselines, including CycleGAN \cite{CycleGAN}, CUT \cite{CUT}, StegoGAN~\cite{StegoGAN}, and EnCo \cite{EnCo}, and diffusion-based baselines, including Pix2Pix-Zero \cite{Pix2Pix-Zero}, CycleNet \cite{xu2023cyclenet}, CycleDiffusion \cite{CycleDiffusion}, SDEdit \cite{SDEdit}, ILVR \cite{ILVR}, EGSDE \cite{EGSDE}, and DiffusionSat \cite{khanna2023diffusionsat}. All trainable baselines use the same FloodScape training split and resolution, while train-free methods use identical pre-disaster inputs and target prompts. DiffusionSat serves as the primary satellite-specific baseline, with the remaining methods providing broader reference comparisons.

\subsubsection{Implementation Details.}

FlowForm adopts an image translation framework similar to InstructPix2Pix~\cite{brooks2023instructpix2pix}. During the diffusion process, the noisy latent representation of the post-disaster image and the latent representation of the pre-disaster image are concatenated along the channel dimension to form the joint input for the U-Net. To use domain-specific priors, we initialize the backbone network with pre-trained weights from DiffusionSat~\cite{khanna2023diffusionsat}. FlowForm was trained on two NVIDIA RTX A6000 GPUs. During training, we utilize AdamW as the optimizer with the learning rate of $2.5 \times 10^{-5}$.

\subsection{Qualitative Evaluation}

\label{sec:experiments}

We present qualitative comparisons between FlowForm and various baseline models.

\begin{figure}[!t]
    \centering
    \includegraphics[width=\columnwidth]{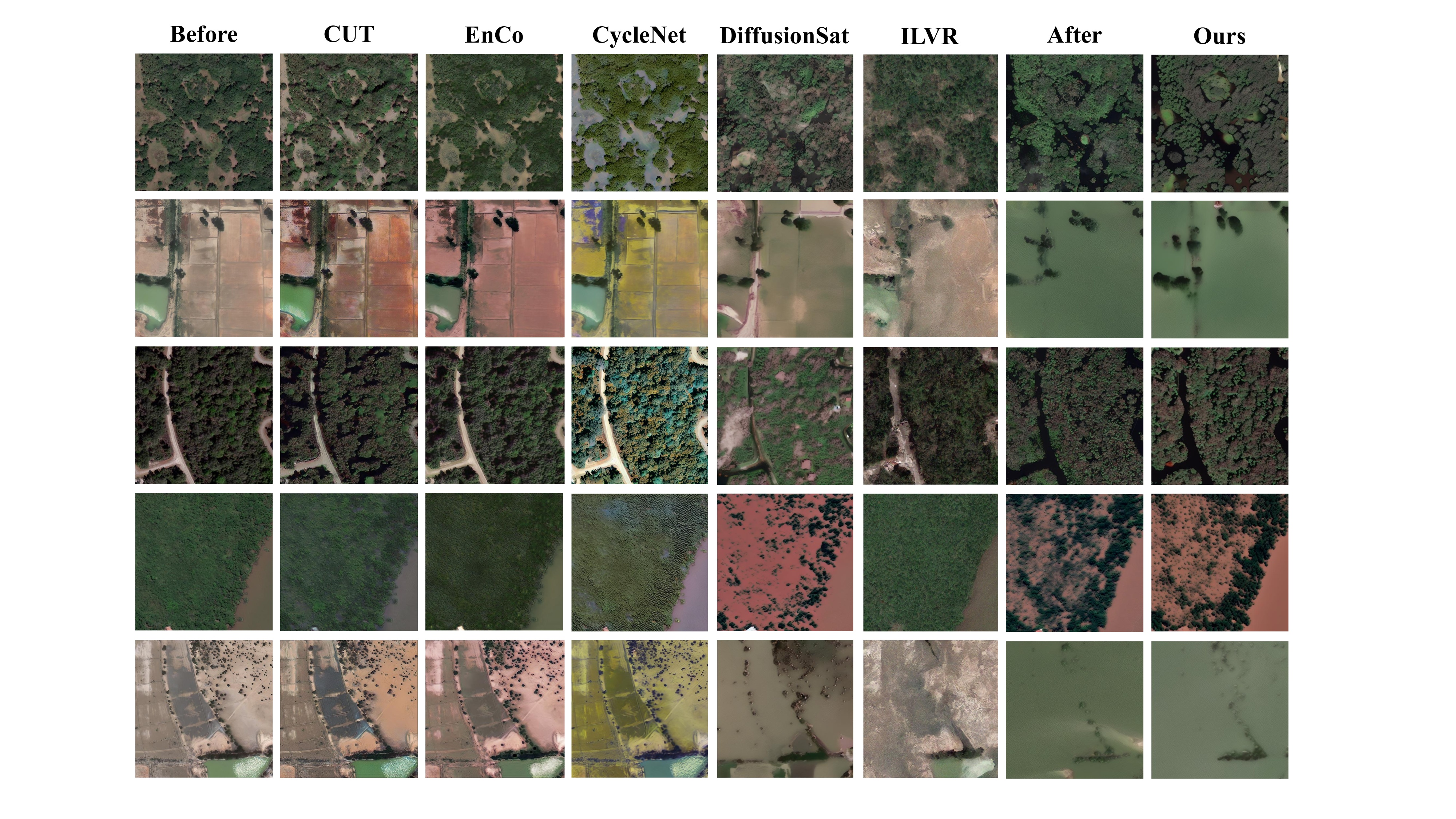}

    \caption{Qualitative comparison on the FloodScape test set.}
    \label{fig:qualitative_comparison}
\end{figure}

Fig.~\ref{fig:qualitative_comparison} compares flood-image generation across several representative baselines. Synthesizing a visually coherent transition from dry to inundated conditions remains challenging. In the displayed examples, I2I models and generic architectures such as CUT, EnCo, and ILVR produce weak or fragmented flood appearance. CycleNet produces layouts resembling flooded regions but introduces visible color shifts that reduce image fidelity.

Among the displayed baselines, DiffusionSat produces coherent water appearance and retains much of the underlying scene structure, but its inundation boundaries are less distinct in these examples. FlowForm shows clearer flood-region delineation while retaining permanent structures and background semantics.

\subsection{Quantitative Evaluation}
\label{sec:quantitative}

We evaluate distributional and semantic image quality with FID~\cite{heusel2017gans} and CLIP~\cite{radford2021learning}, and paired-reference similarity with PSNR~\cite{hore2010image}, SSIM~\cite{wang2004image}, and LPIPS~\cite{zhang2018unreasonable}. Water IoU and FVPS~\cite{lutjens2024generating} measure agreement between the generated and reference flood regions.

As shown in Table~\ref{tab:sota_comparison}, FlowForm achieves the best performance across all reported metrics. Its higher Water IoU indicates stronger agreement between generated and reference flood regions, while the improvements in SSIM and LPIPS show that this gain is accompanied by better paired-image similarity.

\begin{table*}[!t]
    \centering
    \small
    \begin{tabular*}{\textwidth}{@{\extracolsep{\fill}}lccccccc@{}}
        \toprule
        Method & FID $\downarrow$ & SSIM $\uparrow$ & LPIPS $\downarrow$ & CLIP $\uparrow$ & PSNR $\uparrow$ & IoU $\uparrow$ & FVPS $\uparrow$ \\
        \midrule
        Baseline & 85.8043 & 0.4640 & 0.5849 & 0.8584 & 14.8910 & 0.3724 & 0.3926 \\
        w/ TAA & 77.4323 & 0.4867 & 0.5775 & 0.8611 & 15.3644 & 0.4059 & 0.4140 \\
        w/ $\mathcal{L}_{mask}$ & 83.6728 & 0.4782 & 0.5841 & 0.8477 & 15.5326 & 0.4190 & 0.4174 \\
        w/ $\mathcal{L}_{fluid}$ & 82.5228 & 0.4661 & 0.5892 & 0.8545 & 15.5766 & 0.4088 & 0.4098 \\
        w/ FDM & 81.0326 & 0.4939 & 0.5807 & 0.8594 & 15.4803 & 0.3991 & 0.4090 \\
        FlowForm & \textbf{71.7999} & \textbf{0.4941} & \textbf{0.5671} & \textbf{0.8691} & \textbf{15.7556} & \textbf{0.4413} & \textbf{0.4371} \\
        \bottomrule
    \end{tabular*}

    \caption{Granular ablation study on the FloodScape test set.}
    \label{tab:ablation}
\end{table*}

\begin{table*}[!t]
    \centering
    \footnotesize
    \setlength{\tabcolsep}{0.5pt}
    \begin{tabular*}{\textwidth}{@{\extracolsep{\fill}}lccccccc@{}}
        \toprule
        & \multicolumn{5}{c}{\textbf{Image Quality}} & \multicolumn{2}{c}{\textbf{Flood-region Consistency}} \\
        \cmidrule(lr){2-6}\cmidrule(lr){7-8}
        Method & FID $\downarrow$ & SSIM $\uparrow$ & LPIPS $\downarrow$ & CLIP $\uparrow$ & PSNR $\uparrow$ & IoU $\uparrow$ & FVPS $\uparrow$ \\
        \midrule
        \rowcolor{headerblue}\multicolumn{8}{c}{\textbf{GAN-based Methods}} \\
        CycleGAN~\cite{CycleGAN} & 109.1988 & 0.2755 & 0.7066 & 0.8203 & \underline{14.6525} & 0.1836 & 0.2259 \\
        CUT~\cite{CUT} & 116.8511 & 0.2628 & 0.7421 & 0.8273 & 13.8837 & 0.1917 & 0.2199 \\
        StegoGAN~\cite{StegoGAN} & 115.4038 & 0.2372 & 0.6796 & 0.8199 & 14.5507 & 0.1813 & 0.2316 \\
        EnCo~\cite{EnCo} & \underline{108.4521} & 0.2612 & 0.7214 & 0.8105 & 14.5833 & 0.1720 & 0.2127 \\
        \rowcolor{headerblue}\multicolumn{8}{c}{\textbf{Diffusion-based Methods}} \\
        Pix2Pix-Zero~\cite{Pix2Pix-Zero} & 192.4312 & 0.1984 & 0.7688 & 0.7156 & 13.6214 & 0.1845 & 0.2052 \\
        CycleNet~\cite{xu2023cyclenet} & 131.2567 & 0.2345 & 0.7105 & 0.7932 & 13.9542 & 0.1985 & 0.2355 \\
        SDEdit~\cite{SDEdit} & 147.3154 & 0.2217 & 0.7802 & 0.7776 & 13.7937 & 0.1861 & 0.2016 \\
        CycleDiffusion~\cite{CycleDiffusion} & 195.8423 & 0.2514 & 0.6842 & 0.8167 & 14.5211 & 0.1656 & 0.2173 \\
        ILVR~\cite{ILVR} & 187.6626 & 0.1851 & 0.7517 & 0.7779 & 14.5588 & 0.0926 & 0.1349 \\
        EGSDE~\cite{EGSDE} & 119.8169 & 0.2694 & \underline{0.6570} & 0.8374 & 14.4796 & 0.1021 & 0.1574 \\
        DiffusionSat~\cite{khanna2023diffusionsat} & 110.5482 & \underline{0.2838} & 0.6703 & \underline{0.8429} & 13.5489 & \underline{0.2644} & \underline{0.2935} \\
        \midrule
        \textbf{FlowForm} & \textbf{104.7455} & \textbf{0.2909} & \textbf{0.6543} & \textbf{0.8603} & \textbf{14.7841} & \textbf{0.3785} & \textbf{0.3614} \\
        \bottomrule
    \end{tabular*}

    \caption{Zero-shot quantitative comparison on the 442-sample South Africa flood split. The best results are in \textbf{bold}, and the second-best results are \underline{underlined}.}
    \label{tab:ood_metrics}
\end{table*}

\paragraph{Flood-Region Consistency.}

Water IoU measures the overlap between generated inundation regions and SkySense-O reference water masks. FVPS~\cite{lutjens2024generating}, defined as the harmonic mean of IoU and $(1-\mathrm{LPIPS})$, jointly evaluates flood-region agreement and perceptual similarity. FlowForm achieves the best results on both metrics, indicating more accurate flood-region synthesis while maintaining visual quality.

\subsection{Ablation Study}
\label{sec:ablation}

Table~\ref{tab:ablation} details the contributions of the proposed components. The baseline excludes the proposed structural and fluid constraints and serves as the reference configuration. Adding TAA alone restricts the generative space with multimodal structural priors and improves all reported metrics, reducing FID from 85.8043 to 77.4323. FDM regularizes the auxiliary flood states through mask and SWE-inspired constraints; each variant improves Water IoU, while combining both FDM losses raises SSIM from 0.4640 to 0.4939. The complete model combines TAA's anchoring of persistent scene topology with FDM's guidance of flood-region generation, yielding the best overall performance, including an FID of 71.7999 and a Water IoU of 0.4413, and confirming the complementary roles of the two modules.

\subsection{Zero-shot Evaluation on a Held-Out Geographic Event}
\label{sec:ood}

\begin{figure}[!htbp]
    \centering
    \includegraphics[width=0.85\columnwidth]{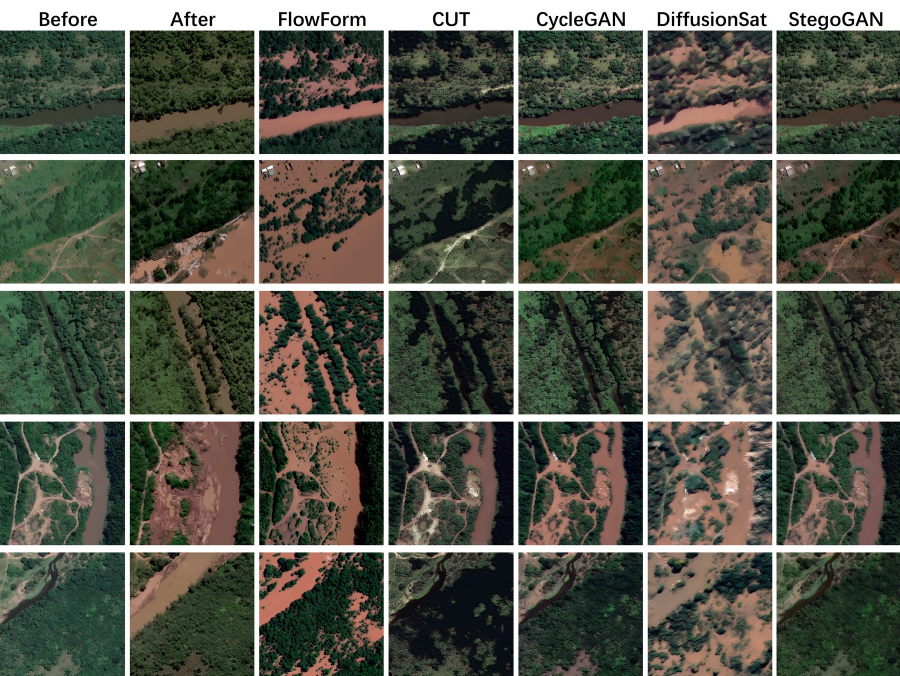}
    \caption{Qualitative zero-shot comparisons on the held-out South Africa event.}
    \label{fig:ood_generalization}
\end{figure}

We evaluate zero-shot performance on 442 images from the 2022 South Africa flood event, which was excluded from training. As shown in Table~\ref{tab:ood_metrics}, FlowForm achieves the best performance across all seven metrics, demonstrating effective generalization to this unseen geographic event. Figure~\ref{fig:ood_generalization} provides qualitative comparisons from the same held-out event. FlowForm produces clear and coherent inundation regions while retaining recognizable vegetation and terrain layouts, consistent with the quantitative results.These results validate its potential for global-scale disaster simulations and zero-shot hazard assessments.

\section{Limitations}
\label{sec:limitations}

FlowForm currently formulates flood synthesis as a static 2D mapping from pre- to post-event imagery, while real floods evolve continuously over time and space. This formulation provides an effective starting point for learning inundation patterns from satellite observations. Building on it, future work could model temporal flood evolution from image sequences or extend the framework to 3D scenes for a richer representation of terrain and water.

\section{Conclusion}
\label{sec:conclusion}

We introduced FlowForm, a latent diffusion framework combining fluid-physics-inspired regularization and structure-aware conditioning for satellite flood-image synthesis. FDM imposes steady-state SWE residuals on auxiliary latent flow fields to guide flood-region generation; TAA injects relative-depth, semantic, and edge cues to preserve pre-event scene layouts. Using our curated FloodScape dataset of $\sim$10,000 high-resolution, spatially aligned pre- and post-disaster satellite image pairs, FlowForm achieves top results across seven metrics on the test set and a held-out South African event. Ablations confirm FDM and TAA's complementary contributions in synthesizing satellite flood images with high visual fidelity, structural consistency, and flood-region agreement.

\bibliographystyle{plainnat}
\bibliography{main}
\end{document}